\documentclass[lettersize,journal]{IEEEtran}
\usepackage{amsmath,amsfonts}
\usepackage{array}
\usepackage[caption=false,font=normalsize,labelfont=sf,textfont=sf]{subfig}
\usepackage{textcomp}
\usepackage{stfloats}
\usepackage{url}
\usepackage{verbatim}
\usepackage{graphicx}
\usepackage{cite}
\usepackage{bm}
\usepackage{color}
\usepackage{booktabs}
\usepackage{multirow}
\usepackage{makecell}
\usepackage{graphicx}
\usepackage{soul}
\usepackage{xcolor}
\definecolor{mygray}{gray}{0.5}
\usepackage{amssymb}
\usepackage[colorlinks=true,linkcolor=red,citecolor=blue,urlcolor=blue]{hyperref}

\usepackage{textcase}
\usepackage[ruled]{algorithm2e}
\usepackage{algpseudocode}

\begin{document}

\title{LLM-attacker: Enhancing Closed-loop Adversarial Scenario Generation for Autonomous Driving with Large Language Models}

\author{Yuewen Mei,
        Tong Nie,
        Jian Sun$^\dagger$,
        Ye Tian,~\IEEEmembership{Senior Member,~IEEE}

\thanks{
*Research supported by the National Natural Science Foundation of China under Grants [52125208, 52422215].

The authors are with the Department of Traffic Engineering and Key Laboratory of Road and Traffic Engineering, Ministry of Education, Tongji University. Shanghai, China. 201804. 
Tong Nie is also with the Department of Civil \& Environmental Engineering, The Hong Kong Polytechnic University.

$^\dagger$Corresponding authors: Jian Sun (sunjian@tongji.edu.cn)
}
}

\maketitle

\begin{abstract}
Ensuring and improving the safety of autonomous driving systems (ADS) is crucial for deployment of highly automated vehicles, especially in safety-critical events.
To address the rarity issue, adversarial scenario generation methods are developed, in which behaviors of traffic participants are manipulated to induce safety-critical events. However, existing methods still face two limitations. 
First, identification of the adversarial participant directly impacts the effectiveness of the generation. 
However, complexity of real-world scenarios, with numerous participants and diverse behaviors, makes identification challenging.
Second, potential of generated safety-critical scenarios to continuously improve ADS performance remains underexplored.
To address these issues, we propose LLM-attacker: a closed-loop adversarial scenario generation framework leveraging large language models (LLMs). Specifically, multiple LLM agents are designed and coordinated to identify optimal attackers. Then, the trajectories of attackers are optimized to generate adversarial scenarios. These scenarios are iteratively refined based on the performance of ADS, forming a feedback loop to improve ADS.
Experimental results show that LLM-attacker can create more dangerous scenarios than other methods, and the ADS trained with it achieves a collision rate half that of training with normal scenarios.
This indicates the ability of LLM-attacker to test and enhance the safety and robustness of ADS.
The framework's closed-loop design enables continuous scenario evolution compliant with regulatory standards, supporting both safety assurance and policy verification for ADS.
Video demonstrations are provided at:
\url{https://drive.google.com/file/d/15rROV_8LUcc2jXSuSNBHVOHMCFKSn__B/view}.

\end{abstract}


\section{Introduction}
\IEEEPARstart{H}{ighly} automated vehicles (HAVs) are advancing rapidly, but the consequences of HAVs accidents can be fatal \cite{scanlon2021waymo}. Therefore, ensuring the safety of autonomous driving systems (ADS) across various scenarios, particularly in safety-critical scenarios prone to accidents, is vital for the development and practical deployment of HAVs. 
Training and testing the safety and robustness of ADS heavily rely on safety-critical scenarios. However, due to their rarity in the real world, limited high-value scenario data are insufficient to fully support both training and testing processes.
Consequently, generating more safety-critical scenarios from real-world traffic data becomes an appealing solution.

To effectively generate safety-critical scenarios, there exist two directions of efforts. 
The first direction focuses on scenes, that is, combinations of scenario parameters such as initial velocity and car-following distance. Specifically, previous work has applied importance sampling techniques \cite{zhao2016accelerated, huang2017accelerated,  yang2023adaptive} and surrogate models \cite{adoe2022sun,zhang2023accelerated,mei2024dbn} to accelerate the determination of safety-critical parameters.
Although these approaches achieve rapid and efficient generation, they are limited to scene-level acceleration and do not fully capture the dynamic behaviors of traffic participants.
On the other hand, the second direction focuses on behaviors. Adversarial scenario generation methods are proposed to generate novel scenarios at the behavioral level, where safety-critical events are commonly achieved by modifying the trajectory of traffic participants \cite{wang2021advsim}. The participant is defined as attacker, inducing aggressive behavior toward the ADS-controlled ego vehicle. This behavior-level manipulation allows for more realistic and generalizable generation of safety-critical scenarios, promising for more effective testing than sampling-based methods \cite{rempe2022generating,cat-zhang-2023,zhang2024module}. 
These scenarios can also support regulatory sandboxes for safety assurance.

However, existing methods for generating adversarial scenarios continue to face two unexplored aspects.
First, since these adversarial methods rely on the behavior of the attacker, one prerequisite of an effective adversarial scenario is to identify proper attackers.
Treating all vehicles as potential attackers is neither efficient nor realistic. 
Typically, there exist a large number of vehicles with complex behaviors in real-world scenarios. 
The interaction relationships between vehicles are inherently nonlinear, with vehicle states such as trajectory and velocity dynamically influenced by the environment and driver behavior \cite{liu2022response}. Furthermore, these behavioral relationships are often embedded within temporal sequences and multivariate features. The nonlinearity and time-variance problems present challenges in identifying the proper number of vehicles, as well as those are potentially aggressive, to serve as attackers. 
In existing work, attackers are typically prescribed by manual design or predefined rules \cite{cat-zhang-2023}, lacking a standardized and adaptive approach applicable to various complex scenarios.

Second, existing methods prioritize the effectiveness of attacks during the testing process, leading to limited integration with the training process. Although these methods improve the attack effectiveness, their practical applicability to enhance the robustness of ADS remains unclear. Specifically, existing methods typically rely on complicated generative models, such as diffusion model \cite{xu2023diffscene} and variational autoencoder \cite{rempe2022generating}. The extremely high computational costs prevent these methods from being incorporated into the online training loop. 
Therefore, the impact of the generated scenarios on the continuous improvement of ADS needs to be explored. It anticipates a simple-yet-effective framework that can ensure high attack effectiveness with a low computational budget.

\begin{figure}[t]
    \centering
    \includegraphics[width=0.8\linewidth]{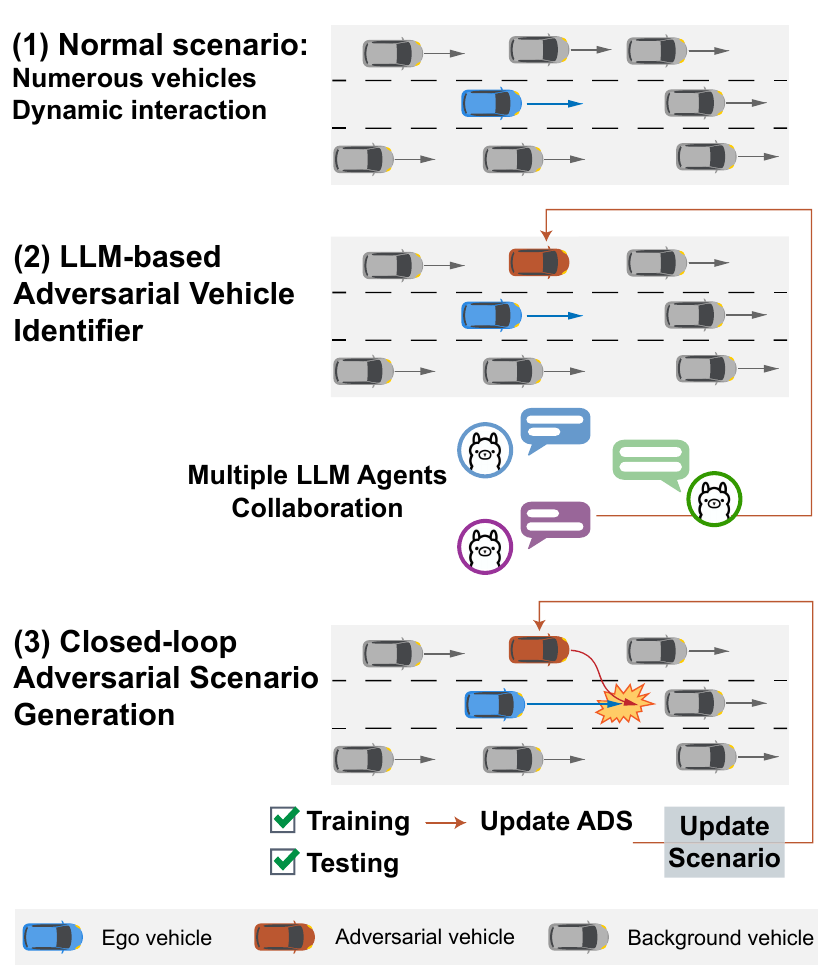}
    \caption{Illustration of LLM-attacker framework. (1) Normal scenario characterized by numerous vehicles and dynamic interactions. (2) LLM-based Adversarial Vehicle Identifier. (3) Closed-loop adversarial scenario generation.} 
    \label{fig:intro}
\end{figure}

The above gaps prompt us to develop an alternative adversarial scenario generation framework.
The emergence of large language models (LLMs) has attracted considerable attention due to their powerful context reasoning and understanding capabilities \cite{touvron2023llama, yu2024llm}. 
Encouragingly, LLMs have demonstrated potential in understanding traffic situations and vehicle behaviors in the context of ADS perception tasks \cite{yang-2024-llm4drivesurveylargelanguage, marcu2024lingoqa}. 
Based on their world knowledge, LLMs can possess a profound understanding of complex traffic scenarios, especially multi-object  interactions between the ego vehicle and background vehicles.
This provides a new perspective for solving the aforementioned limitations.
Consequently, we propose a closed-loop adversarial scenario generation framework utilizing LLMs, termed LLM-attacker. 
Specifically, suitable prompt engineering is designed to elicit their scenario understanding capabilities in distinguishing interaction relationship. Then, multiple LLM agents are coordinated to identify suitable attackers within complex scenarios. 
The identified attackers and their adversarial trajectories, which compose the generated adversarial scenarios, are then used to test and train the ADS. 
During the training process, the adversarial scenarios are continuously updated based on the performance of the trained ADS, forming the closed-loop adversarial scenario generation.
Our contributions can be summarized as follows.
\begin{itemize}
\item{
A collaborative LLM-based agentic framework is designed to automatically identify optimal attackers.}
\item{A closed-loop adversarial scenario generation framework using LLMs with adversarial training is developed. }
\item{The effectiveness of safety-critical scenarios is validated on both ADS safety testing and training, which achieve the evolution of scenarios and ADS simultaneously.}
\end{itemize}

Rest of this paper is organized as follows. 
Section \ref{sec:related work} briefly reviews related work.
Section \ref{sec:methodology} formulates the methodology.
Section \ref{sec:experiment} conducts experiments to verify effectiveness of the proposed methodology. Section \ref{sec:conclusion} concludes the work.

\section{Related Work}\label{sec:related work}
\subsection{Safety-critical scenarios generation of ADS}
Safety-critical scenarios, which result in severe consequences, are extremely rare in real world. Consequently, recent work increasingly focus on generating more safety-critical scenarios for effective ADS safety issue\cite{ding2023survey}. Generation methods can be categorized into knowledge-based \cite{L2C2020ding}, acceleration-based \cite{zhang2023accelerated,YANG2025107962,mei2024high}, and adversarial-based \cite{wang2021advsim, rempe2022generating}. Among these methods, adversarial-based methods are characterized by efficiency and authenticity. In our work, it is expected that the generated adversarial scenarios can be further exploited for ADS safety from the perspective of testing and training.  

\subsection{Adversarial training for ADS}
Adversarial training is a crucial method for enhancing the robustness of ADS, especially in the aspect of object detection. 
Only a few works focus on adversarial training of end-to-end driving or decision-making tasks of ADS \cite{cat-zhang-2023,zhang2024module}.
Zhang et al. \cite{cat-zhang-2023} propose a closed-loop adversarial training framework to improve the safety performance of end-to-end ADS. The safety-critical scenarios with an attacker are resampled for training of the ADS policy. Existing frameworks only focus on the single and fixed attacker during the training process. However, complex scenarios with multi-attackers are even more dangerous for ADS. Meanwhile, the most-efficient attackers vary between different situations, which should be adaptive and automatic adjusted in different scenarios. 

\subsection{LLMs for ADS}
LLMs have been used for the development of end-to-end ADS in various aspects \cite{yang-2024-llm4drivesurveylargelanguage}, such as improving the adaptability of the perception system, enhancing the planning ability of ADS, generating realistic scenario videos. LLMs also show great potential to construct traffic scenarios for ADS. For example, ChatScene \cite{chatscene_2024_CVPR} converts natural language descriptions of traffic scenarios into practical simulations by building a retrieval database. Meanwhile, LLMs are also applied to generate safety-critical scenarios. CRITICAL \cite{tian2024_critical} modifies the parameters of the highway scenario by LLM to make scenarios more challenging for ADS, demonstrating the ability of LLMs to understand the scenario. Our work aims to find an effective method to exploit the reasoning capability of LLMs to solve the vehicle identification problem in the generation of adversarial scenarios.

\section{Methodology}\label{sec:methodology}
In this section, we proposed LLM-attacker: the closed-loop adversarial scenario generation framework leveraging LLMs. First, preliminaries of adversarial attack based on Reinforcement Learning (RL) is introduced. Second, the LLM-based Adversarial Vehicle Identifier, which is built on the collaborations of multiple LLM agents, is designed. Then, we elaborate on the practical implementation of LLM-attacker.

\subsection{Overview}  
Autonomous driving task based on RL can be formulated as a Markov Decision Process (MDP), which is defined by the tuple $(\mathcal{S}, \mathcal{A}, R, f)$. $\mathcal{S}$ represents the state space, composed of vehicle state, navigation information, and lidar point clouds. $\mathcal{A}$ stands for the action space that contains the acceleration and steering commands of the autonomous vehicle. $R$ is the reward function defined by the completion rate and safety status of driving task. The dynamic variation of the state with the action is defined by the transition function $f$. To perform the autonomous driving task safely, RL is required to maximize the expected return $J(\pi,f)$:

\begin{equation}
J(\pi,f) = \mathbb{E}_{a_t\sim\pi(s_t), s_{t+1}\sim f(s_{t},a_{t})}\left[\sum_{t=0}^T R(s_t, a_t)\right],
\end{equation}
where $\pi$ is the policy learned by RL. $T$ is the duration of the traffic scenario.

The adversarial attack $f^{adv}$ aims to disturb autonomous driving tasks by intentionally introducing background vehicles with adversarial behaviors, increasing the risk of failure and decreasing the reward. The target of adversarial attack is: 
\begin{equation}
\min\limits_{f^{adv}} J(\pi,f^{adv}).
\end{equation}

Then, adversarial training aims to improve the ability of ADS to handle unexpected adversarial situations, ultimately enhancing the robustness of the RL policy under $f^{adv}$.
\begin{equation}
\max\limits_{\pi}\min\limits_{f^{adv}} J(\pi,f^{adv}).
\end{equation}

In each scenario, $J(\pi,f)$ equals the lowest value when RL fails on the driving task, i.e., collides with other obstacles. 
Suppose that the adversarial attack starts from the moment $t$, and current state $ X = (W, S_{1:t}^{ego}, \bm{\mathit{S}}_{1:t}^{bac})$ is provided. $W$ stands for the road geometry of the scenario. $S_{1:t}^{ego}$ is the trajectory of the ego vehicle and $\bm{\mathit{S}}_{1:t}^{bac}$ is the set of trajectories of the $M$ background vehicles from initial time to moment $t$.  Trajectories of the ego vehicle and the $N$ adversarial vehicles in the future are denoted as $Y^{ego} = S_{t:T}^{ego}$ and $\bm{\mathit{Y}}^{adv} = \bm{\mathit{S}}_{t:T}^{adv}$ respectively.
Consequently, the adversarial attack can be converted to find the optimal $\bm{\mathit{Y}}^{adv}$ to maximize the posterior probability of collision between $\bm{\mathit{Y}}^{adv}$ and $Y^{ego}$ based on $X$:

\begin{equation}
    \min\limits_{f^{adv}} J(\pi,f^{adv}) \Leftrightarrow \max\limits_{\bm{\mathit{Y}}^{adv}} \mathbb{P}(\bm{\mathit{Y}}^{adv}, Y^{ego} \mid X, C ),
\end{equation}
where $C$ represents the occurrence of collisions.

The optimization of $\bm{\mathit{Y}}^{adv}$ can be factorized into two steps. First, identifying the optimal $N$ adversarial vehicles from the $M$ background vehicles, which is performed by the LLM-based adversarial vehicle identifier (Section III-B). Second, calculating the adversarial trajectories of the $N$ adversarial vehicles (Section III-C).

\subsection{LLM-based Adversarial Vehicle Identifier}
\begin{figure}
    \centering
    \includegraphics[width=0.98\linewidth]{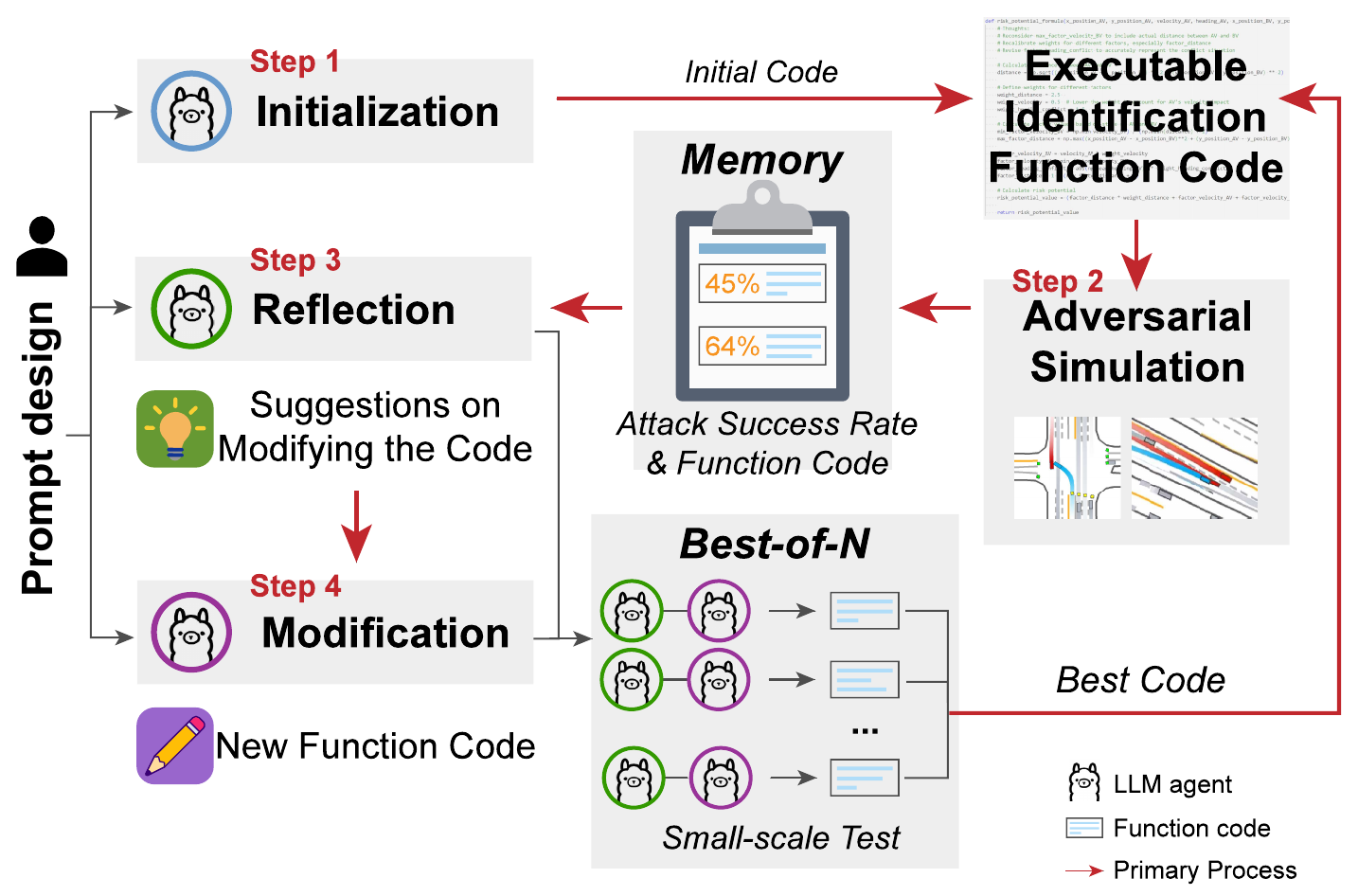}
    \caption{Agentic Framework of the LLM-based Adversarial Vehicle Identifier.}
    \label{fig:method_llm}
\end{figure}
The first step to generate adversarial scenarios is the identification of attackers. Each scenario contains a large number of background vehicles, and the scenarios faced by autonomous driving are numerous. Manually assigning the adversarial vehicles that have the greatest impact on the ego vehicle in each scenario would be labor-intensive and time-consuming. Therefore, the LLM-based Adversarial Vehicle Identifier is proposed to achieve automatic and efficient identification.

The agentic framework of the LLM-based Adversarial Vehicle Identifier is shown in Fig. \ref{fig:method_llm}. There are three LLM-based modules, including \texttt{Initialization}, \texttt{Reflection}, and \texttt{Modification}. First, the \texttt{Initialization} module is designed to generate the initial form of the identification function code. Second, the adversarial vehicles in each scenario are identified based on the function code. The adversarial attack is conducted in simulation to verify the attack success rate. Third, the generated code and the attack success rate are given to the \texttt{Reflection} module, which outputs specific analysis and modification suggestions. Last, the \texttt{Modification} module generates a new function code based on the suggestions. The simulation, \texttt{Reflection} and \texttt{Modification} process, i.e., step 2 to step 4, is repeated for $Z$ iterations. Then, a collaborative multi-agent framework is formed, where each LLM-based module serves a specialized role and contributes to a coordinated workflow. Collaboration is carried out through the structured information flow. Each module receives the previous output, processes it according to its task-specific prompt, and passes on the enhanced result, enabling an iterative refinement process across multiple iterations.

Since LLM relies on textual reasoning, skillful prompt design can help LLM better understand the task and improve the effectiveness of generation. The details of the prompt engineering for the three LLM-based modules are as follows. 

\subsubsection{Initialization Module}
The \texttt{Initialization} module is responsible for generating the initial executable identification function code. The instructions provided for this module include role setting, background information, task description, input parameters of the function, requirements for the function, and output format. The input parameters include the state of the ego vehicle and the background vehicles. The \texttt{Initialization} module is denoted as:
\begin{equation}
    F_0 = \mathrm{FunctionInit}(Task, Requi),
\end{equation}
where $F_0$ is the initial function code, $\mathrm{FunctionInit(\cdot)}$ stands for the invocation of \texttt{Initialization} module, $Task$ and $Requi$ refer to the task and requirements of \texttt{Initialization} Module respectively.

The \textbf{chain-of-thought} (CoT) prompting strategy can help LLM understand and solve the problem step by step \cite{wei2022chain}. Therefore, CoT is applied in the prompt. For instance, in the task description, the function code generation task is divided into three reasoning steps: a) identifying the factors that should be considered in the function, b) considering how these factors change over time, and c) ensuring that the weighting and normalization of each item are accounted for. The CoT instructions clearly state the elements to be concerned with at each step and guide LLM reason progressively. In a similar manner, the CoT prompting is also designed in the \texttt{Reflection} module and \texttt{Modification} module.

\subsubsection{Reflection Module}
Based on the generated the function code, the adversarial vehicles can be identified and attack on ADS can be conducted in the simulation. After verifying the attack success rate, it is necessary to analyze and improve the current function. Therefore, the \texttt{Reflection} module is designed for analyzing the reasonableness of current function and giving specific suggestions on modification. 
The instructions provided for this module include role setting, background information, task description, current function code, current success rate, analysis requirements, and output requirements. The \texttt{Reflection} module and Simulation is denoted as:
\begin{equation}
\begin{aligned}
A_i = \mathrm{Function}&\mathrm{Refl}(Task, Requi, Memo, F_{i-1}, E_{i-1}), \\
&E_i = \mathrm{Simulation}(F_i),
\end{aligned}
\end{equation}
where $A_i$ is the specific suggestions on modification, $\mathrm{FunctionRefl(\cdot)}$ represents the invocation of \texttt{Reflection} module, $Memo$ refers to memory, $F_{i-1}$ and $E_{i-1}$ are the code and evaluation result of current function, $\mathrm{Simulation(\cdot)}$ refers to conduct simulation test and obtain the attack success rate based on $F_i$.

After the first iteration, both the previous function and its corresponding attack success rate are fed as \textbf{Memory} to the module. The \texttt{Reflection} module is guided to observe and learn from the differences between previous and current functions. Meanwhile, the attack success rate serves as the evaluation feedback to the suggestion made by the \texttt{Reflection} module, helping it learn how to make more appropriate suggestions. The Memory mechanism directs the \texttt{Reflection} module to gradually improve toward an excellent analyzer and suggester in adversarial vehicle identification over successive iterations.

\subsubsection{Modification Module}
The \texttt{Modification} module modifies the current function according to the suggestion given by the \texttt{Reflection} module. Therefore, role setting, input parameters of the function, current function, suggestions on modification, requirements for the function, and output format are provided for this module to generate an improved executable identification function code. The \texttt{Modification} module is denoted as:
\begin{equation}
    F_{i+1} = \mathrm{FunctionModi}(Task, Requi, F_i, A_i),
\end{equation}
where $F_{i+1}$ is the function code generated in iteration $i$, $\mathrm{FunctionModi(\cdot)}$ represents the invocation of \texttt{Modification} module.

\begin{algorithm}[t]
\label{algorithm_llm}
\caption{{LLM-based Automatic Adversarial Vehicle Identifier}}
\begin{small}
\begin{algorithmic}[1]
\Require
Number of iteration $Z$, Number of candidates in each iteration $Q$, Initial small-scale simulation test result $e_0$
    \State $\triangleright$ \textit{Initialize executable identification function code}
    \State $F_0 = \mathrm{FunctionInit}(Task, Requi)$
    \State $E_0 = \mathrm{Simulation}(F_0)$     
    \State $\triangleright$ \textit{Improve executable identification function code}
    \State \textbf{for} $i=1,\dots,Z$ \textbf{do}
    \State \hspace{0.2cm} \textbf{for} $j=1,\dots,Q$ \textbf{do}   $\quad  \triangleright$ \textit{Best-of-N}
    \State \hspace{0.4cm}  $A_{ij} = \mathrm{FunctionRefl}(Task, Requi, Memo, F_{i-1}, E_{i-1})$
    \State \hspace{0.4cm} $F_{ij} = \mathrm{FunctionModi}(Task, Requi, F_{i-1}, A_{ij})$
    \State \hspace{0.4cm} $e_{ij} = \mathrm{TestSim}(F_{ij})$
    \State \hspace{0.4cm} \textbf{if} $e_{ij}>e_{i-1}$ \textbf{then}
    \State \hspace{0.6cm} $F_i = F_{ij}$
    \State \hspace{0.6cm} \textbf{break}
    \State \hspace{0.4cm} \textbf{elif} $j = Q$ \textbf{then}
    \State \hspace{0.6cm} $F_i = \arg\max_{e_{ij}} F_{ij}$
    \State \hspace{0.4cm} \textbf{end if}
    \State \hspace{0.2cm} \textbf{end for}
    \State \hspace{0.2cm} $E_i = \mathrm{Simulation}(F_i)$
    \State \textbf{end for}
    \State \textbf{Return} $F_Z$
\end{algorithmic}
\end{small}
\end{algorithm}

Even with the same input, LLM can produce outputs with significant differences, both in terms of content and quality. To address the inherent stochasticity in LLMs and improve the quality of generation, the \textbf{Best-of-N} sampling strategy \cite{wang2023SelfConsistency} is adopted. Specifically, in each iteration, the \texttt{Reflection} and \texttt{Modification} modules are invoked multiple times to generate up to $Q$ different function code candidates. A small-scale simulation test is then performed to evaluate the attack success rate for each candidate. If the test result of any candidate surpasses the current function's performance, that candidate is selected as the new function code for the next iteration. Otherwise, when the number of generation is equal to $Q$, the best performing candidate among $Q$ is chosen and output as the final function code. The small-scale simulation test is denoted as:
\begin{equation}
e_{ij} = \mathrm{TestSim}(F_{ij}),
\end{equation}
where $e_{ij}$ is the small-scale test result of $F_{ij}$, $\mathrm{TestSim}$ represents conduct small-scale simulation test, $F_{ij}$ is the $j$th function code generated in iteration $i$.

Large modifications made by LLM to the function can lead to drastic decreases, even when the attack success rate has already reached a high level. To avoid such instability, a \textbf{Local Optimization} approach is employed in each iteration when the attack success rate of the current function exceeds a predefined threshold. In particular, the \texttt{Reflection} module is firstly instructed to suggest only minor adjustments, including modifying the weights of an item in the function, without modifying the structure of the function. If minor adjustments fail to make improvement after $Q/2$ modification, the \texttt{Reflection} module is then allowed to propose structural modifications to explore the potential for better performance. In this way, it is achieved that the function can be upgraded through both controlled refinement and exploration of new structures, while maintaining its current performance.

\begin{figure*}
    \centering
    \includegraphics[width=0.95\linewidth]{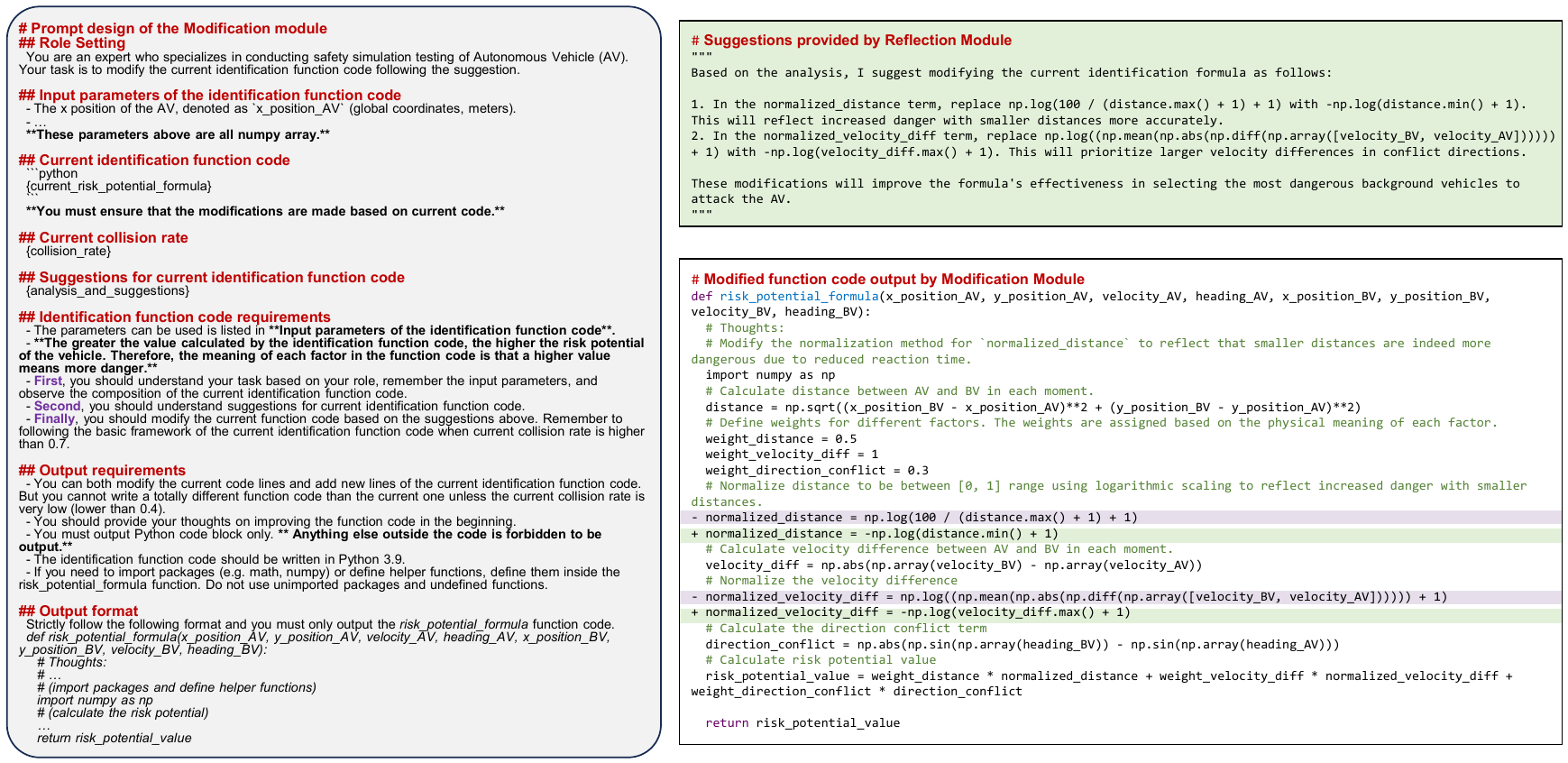}
    \caption{The left side of the figure presents the prompt design of the \texttt{Modification} module, while the upper right side shows the suggestions provided by the \texttt{Reflection} module, and the lower right side displays the modified code made by the \texttt{Modification} module.}
    \label{fig:prompt}
\end{figure*}

In summary, the implementation of the LLM-based Adversarial Vehicle Identifier is shown in Algorithm \eqref{algorithm_llm}. The identification of $N$ adversarial vehicles from $M$ background vehicles is denoted as: $N \Leftarrow F_Z(M)$. We take the prompt of the \texttt{Modification} module as an example, shown in Fig. \ref{fig:prompt}, containing the role setting, suggestions, requirements, and so on. A concrete example of the suggestions provided by the Reflection module and the corresponding modified code made by the \texttt{Modification} module is also demonstrated in Figure \ref{fig:prompt}.

\subsection{Closed-loop Trajectory Generation for Multiple Attackers}
The second step of adversarial scenario generation is to compute the trajectory of multiple adversarial vehicles $\bm{\mathit{Y}}^{adv}$.
It is assumed that the state of vehicles in the future are independent with each other. 
Meanwhile, the collision merely depends on the trajectories of the ego vehicle and the adversarial vehicles. Therefore, the collision probability to be maximized can be factorized to a tractable form \cite{cat-zhang-2023} based on the Bayes theorem under this assumption.

\begin{equation}\label{bayes}
\begin{aligned}
& \max\limits_{\bm{\mathit{Y}}^{adv}} \mathbb{P}(\bm{\mathit{Y}}^{adv}, Y^{ego} \mid X, C) \propto \\
& \max\limits_{\bm{\mathit{Y}}^{adv}} \mathbb{P}(\bm{\mathit{Y}}^{adv}\mid X) \mathbb{P}(Y^{ego}\mid X) \mathbb{P}(C\mid Y^{ego},\bm{\mathit{Y}}^{adv}) \propto \\
& \max\limits_{\bm{\mathit{Y}}^{adv}}\prod_{i=1}^{N} \mathbb{P}(\mathit{Y}_i^{adv}\mid X) \mathbb{P}(Y^{ego}\mid X) \mathbb{P}(C\mid Y^{ego},\mathit{Y}_i^{adv})
\end{aligned}
\end{equation}
where candidate trajectories of $N$ adversarial vehicles $\bm{\mathit{Y}}^{adv}$ and the corresponding probability $\mathbb{P}(\bm{\mathit{Y}}^{adv}\mid X)$ are given by a pre-trained probabilistic traffic forecasting model DenseTNT \cite{densetnt_2021_ICCV}. The trajectory of ego vehicle $Y^{ego}$ and probability $\mathbb{P}(Y^{ego}\mid X)$ are obtained from the rollouts of ego vehicle. Based on the trajectory of adversarial vehicle and ego vehicle, the collision likelihood $\mathbb{P}(C\mid Y^{ego},\bm{\mathit{Y}}^{adv})$ can be calculated. 

Based on the calculation results, the adversarial vehicles follow the trajectories with the highest collision likelihood, which turns the normal scenario into the adversarial scenario. The $N$ adversarial vehicles collectively form a multi-vehicle attack against the ADS. The generated adversarial scenarios are applied to both testing and training of ADS.

The pseudo code of LLM-attacker is summarized in Algorithm \eqref{algorithm_overall}. Given a collection of normal scenarios $E$ for training and $E'$ for testing, the attacker identification function $F_z$ is first applied to each scenario to identify the optimal $N$ attackers.
Then, during training process, the traffic forecasting model $\mathcal{K}$ generates candidate adversarial trajectories $\bm{Y}^{adv}$. The ego trajectory $Y^{ego}$ is generated from the ego trajectory buffer $\mathcal{D}$, which changes as the learning of the ADS policy $\pi$ continues. As $\pi$ evolves, the adversarial trajectories $\bm{Y}^{adv}$ also change accordingly, forming a closed-loop adversarial training process.
During testing process, $\bm{\mathit{Y}}^{adv}$ is also calculated based on the performance of ADS. The ADS policy $\pi^*$ remains unchanged. The testing evaluation metrics $\epsilon$ are calculated by testing the ADS in all normal scenarios and generated adversarial scenarios.

\begin{algorithm}
\label{algorithm_overall}
\caption{Closed-loop LLM-attacker framework}
\begin{small}
\begin{algorithmic}[1]
\Require
\parbox[t]{\dimexpr\linewidth-\algorithmicindent}{ADS policy $\pi$, Normal scenarios for training $E$, Normal scenarios for testing $E'$, traffic forecasting model $\mathcal{K}$, learning algorithm $\mathcal{F}$, ego trajectory buffer $\mathcal{D}$, max policy training steps $K_{max}$, trained policy $\pi^*$, testing evaluation metrics $\epsilon$
}
\vspace{0.1cm}
\State $\triangleright$ \textit{Preprocess: Identify adversarial vehicles for each scenario}
\State \textbf{for} each scenario $e_i$ in $E$ and $E'$ \textbf{do}
\State \hspace{0.3cm} $N \Leftarrow F_Z(M)$
\State \textbf{----- Training process -----}
\State \textbf{while} training steps is smaller than $K_{max}$:
\State \hspace{0.1cm} Sample one normal scenario $e$  from $E$
\State \hspace{0.1cm} $\triangleright$ \textit{Calculate adversarial trajectories}
\State \hspace{0.4cm} $\{\bm{\mathit{Y}}^{adv},\mathbb{P}(\bm{\mathit{Y}}^{adv}\mid X)\} \Leftarrow \mathcal{K}(X)$
\State \hspace{0.4cm} $\{ Y^{ego}, \mathbb{P}(Y^{ego}\mid X) \} \Leftarrow \mathcal{D}(X)$
\State \hspace{0.4cm} $\bm{\mathit{Y}}^{adv} \Leftarrow \mathbb{P}(\bm{\mathit{Y}}^{adv}, Y^{ego} \mid X, C)$
\State \hspace{0.1cm} Conduct adversarial simulation and update ego trajectory
\State \hspace{0.1cm} $\mathcal{D} \Leftarrow \mathcal{D} \cup Y^{ego}$
\State \hspace{0.1cm} $\pi \Leftarrow \mathcal{F}(\pi)$
\State \textbf{----- Testing process -----}
\State \textbf{for} each scenario $e_i$ in $E'$ \textbf{do}
\State \hspace{0.1cm} $\triangleright$ \textit{Calculate adversarial trajectories}
\State \hspace{0.4cm} $\{\bm{\mathit{Y}}^{adv},\mathbb{P}(\bm{\mathit{Y}}^{adv}\mid X)\} \Leftarrow \mathcal{K}(X)$
\State \hspace{0.4cm} $\{ Y^{ego}\} \Leftarrow \pi^{\star}(X)$
\State \hspace{0.4cm} $\bm{\mathit{Y}}^{adv} \Leftarrow \mathbb{P}(\bm{\mathit{Y}}^{adv}, Y^{ego} \mid X, C)$
\State \hspace{0.1cm} Conduct adversarial simulation and record $\epsilon$

\State \textbf{Return} Final policy $\pi^{\star}$, evaluation metrics $\epsilon$

\end{algorithmic}
\end{small}
\end{algorithm}



\section{Experiment}\label{sec:experiment}
This section offers experiments to verify the effectiveness of the proposed LLM-attacker framework. We first describe the experimental setup, followed by an analysis of the outputs generated by the LLM. We then test adversarial attack success rate and evaluate effectiveness of adversarial training.
\subsection{Experimental settings}
The autonomous driving task is performed based on the waymo open motion dataset (WOMD), which is a collection of diverse traffic scenarios. 485 complex traffic scenarios were imported from WOMD \cite{Ettinger_2021_ICCV_womd}, covering a wide range of road structures such as straight road, ramp, T-intersection, and intersection. The lightweight and flexible driving simulator called MetaDrive \cite{metadrive_2023} is applied for the experiments. The end-to-end driving policy is trained by a vanilla RL algorithm twin delayed deep deterministic (TD3) \cite{td3-2018}. The observation of RL agent contains ego states, navigation information and surrounding information provided by 2D LiDAR sensor. The output action includes steering, throttle, and brake. The LLM we employed is LLaMA \cite{touvron2023llama} with version number 3.1 and 8 billion parameters, which has demonstrated impressive capabilities in various spatio-temporal reasoning tasks \cite{he2024geolocation}. All experiments are implemented with an NVIDIA Quadro RTX 4000 GPU and Intel Core i7-12700K CPU.

The parameters in the LLM-based Adversarial Vehicle Identifier are set as follows. The number of adversarial vehicles $N$ is set to $1$ or $2$ for the comparison of the effects of the single-vehicle adversarial attack and multi-vehicle adversarial attack. The number of iterations $Z$ and candidates in each iteration $Q$ are set to $8$ and $11$, respectively. The initial result of the small-scale simulation test $e_0$ is assigned to $0.4$. 

\subsection{LLM outputs}

\begin{figure}[h]
    \centering
    \includegraphics[width=0.6\linewidth]{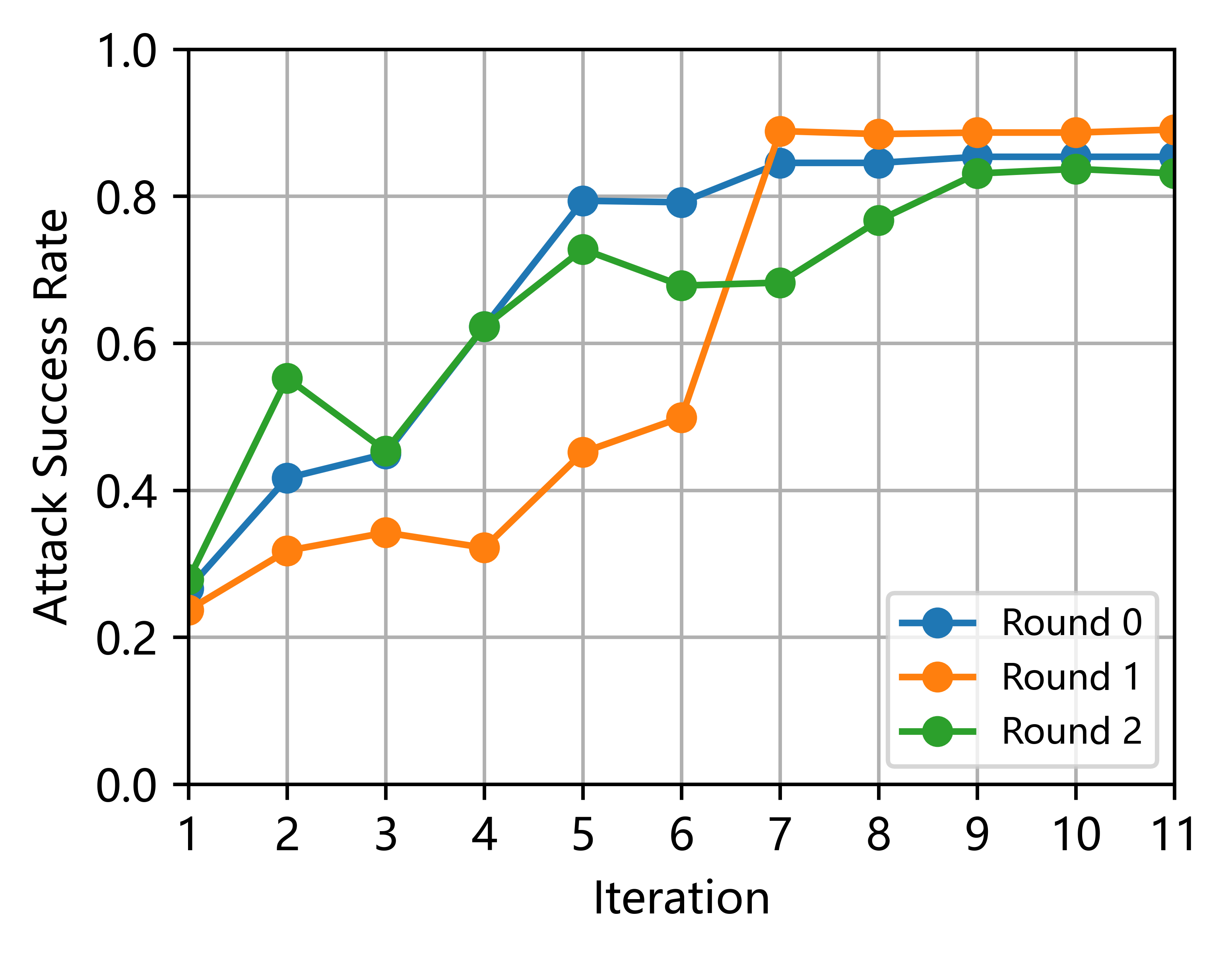}
    \caption{The attack success rate range with the iteration of identification function in the LLM-based Adversarial Vehicle Identifier.}
    \label{fig:LLM_generation_attack_results}
\end{figure}

We perform the LLM-based Adversarial Vehicle Identifier three times and record the attack success rate in each iteration, which is shown in Fig.\ref{fig:LLM_generation_attack_results}. 
It can be seen that the attack success rate increases overall with the iteration of identification function, which demonstrates the continuous improvement effect of LLM. The three rounds of identification end with at least $83\%$ success rate. The best performance is achieved in the 11th iteration of Round 1 and the corresponding function code is used in the subsequent experiments. The average time cost for the three rounds is 4.42 hours, with LLM reasoning taking 2.10 hours and simulation test to validate the attack success rate taking 2.32 hours. 
To ensure full control of the multi-agent prompting pipeline, we run LLaMA 3.1-8B locally. If commercial APIs are used instead, the inference latency can be significantly reduced.

\subsection{Performance evaluation of adversarial vehicle identifier}
In this subsection, we evaluate the LLM-based Adversarial Vehicle Identifier by comparing the adversarial scenarios generated based on different identification methods.
Three other identification methods are compared: (1) Random: adversarial vehicles are randomly identified from all background vehicles. (2) Min TTC: adversarial vehicles are the vehicles having the minimum time to collision (TTC) to ego vehicle. (3) Kinetic Field: The kinetic field represents the influences of the moving background vehicle on driving safety, which is a component of the driving safety field \cite{wang2015driving}. Two kinds of agent are tested as the ego vehicle: (1) Log agent: Replay of the real-world logged ego vehicle trajectories. (2) RL agent: Pretrained RL driving policy, which is trained by real-world logged scenarios. 

Scenarios generated from different identification methods are evaluated from two aspects: the effectiveness of attack and the realism of driving behavior. The effectiveness is quantified by the collision ratio of ego vehicle across all generated scenarios, i.e. the attack success rate of adversarial vehicles. Higher rates indicate more effective attacks. To assess realism, we measure Kullback–Leibler (KL) divergence between attacker’s acceleration distributions in generated and real-world scenarios. The rate of abnormal jerk is also calculated. Lower values of these metrics represent higher levels of realism.

In terms of effectiveness, the adversarial attack success rate is shown in TABLE \ref{tab:collision rate}. The LLM-attackers achieve the highest success rate in all experiments. As for realism, the evaluation results are shown in TABLE \ref{tab:realism}. The occurrence of abnormal jerk rate in LLM-attacker is the lowest, with a relatively low acceleration KL divergence, demonstrating the realism of attacker's behavior.

To verify effect of the number of adversarial vehicles $N$, experiments are conducted to compare one, two, three, four, and five attackers. Average computational cost for each scenario is also recorded. It can be concluded from TABLE \ref{tab:collision rate} that: (1) Scenarios with two attackers achieve a significantly higher attack success rate compared to scenarios with a single attacker. (2) When the number of attackers increases to three or four, improvement in attack success rate becomes marginal. 
While multiple methods yield the same success rate of 54.79\%, collisions happen in different scenarios, reflecting distinct attack behaviors across different methods.
(3) The computation time increases with the number of attackers. Therefore, considering both attack effectiveness and computational efficiency, two attackers yield the best overall performance.

\begin{figure}[!htbp]
    \centering
    \includegraphics[width=0.93\linewidth]{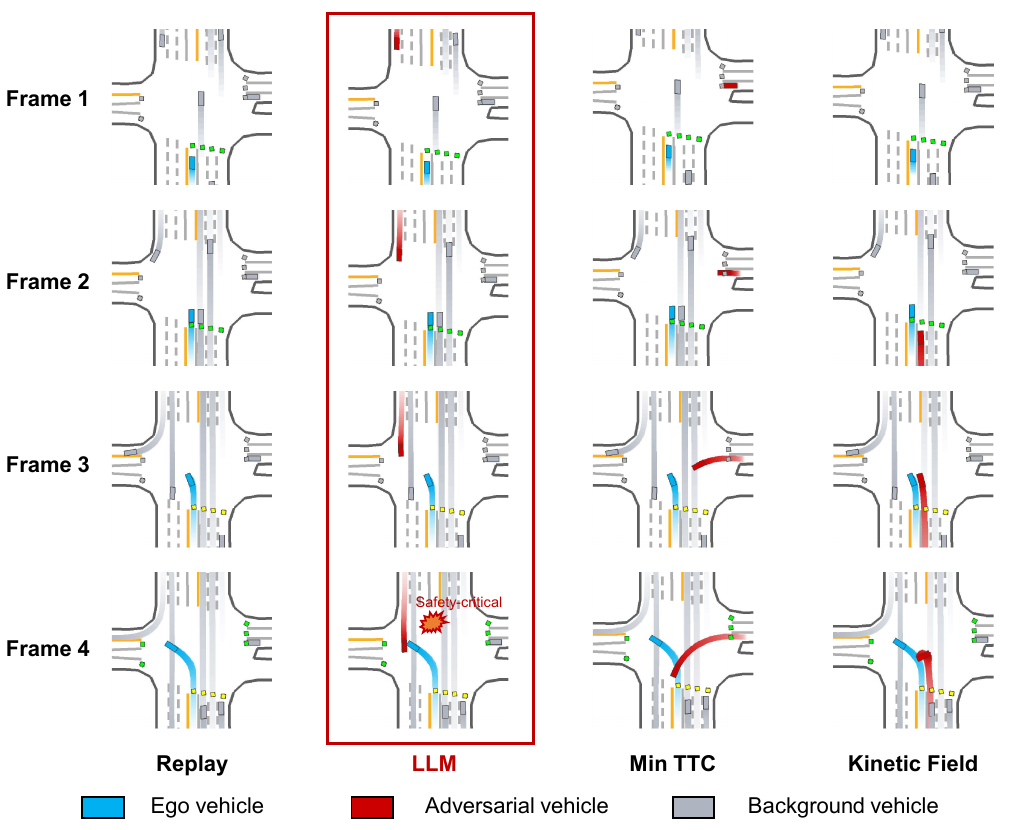}
    \caption{An identification and safety-critical scenario example. The blue rectangle is the ego vehicle controlled by ADS and the red rectangle is the identified attacker. Four frames are sampled to represent the dynamics of the entire scenario. The real-world replay scenario, the attacker identification and generated adversarial scenarios by LLM, Min TTC and Kinetic Field are displayed from left to right. LLM successfully converts real-world normal scenarios into safety-critical scenarios, whereas Min TTC and Kinetic Field fail to achieve so.}
    \label{fig:identification_example}
\end{figure}

\begin{table*}[!htbp]
    \fontsize{9pt}{13pt}\selectfont
    \vspace{-10pt}
    \centering
    \caption{Adversarial Attack Success Rate and Average Computational Cost}
    \begin{tabular}{c|cc|cc|cc|cc|cc}
    \hline
    \multirow{2}{*}{Identification Method} &  \multicolumn{2}{c}{One attacker} & \multicolumn{2}{c}{Two attackers} &  \multicolumn{2}{c}{Three attackers} & \multicolumn{2}{c}{Four attackers} & \multicolumn{2}{c}{Five attackers}\\
    & Log & RL & Log & RL & Log & RL & Log & RL & Log & RL \\
    \hline
    Random &  18.13\% &22.08\% & 38.96\% & 26.88\% & 51.67\%& 33.3\% & 59.17\%& 36.88\% & 63.75\%&36.88\%\\  
    Min TTC & 54.17\%  & 37.92\% & 76.25\% & 45.42\% &82.29\% &50.21\% &85.42\% &53.96\%  & 87.50\%&53.96\%\\
    Kinetic Field & 57.71\% & 37.50\%  & 83.13\% & 44.58\% & 88.13\%&47.08\% &89.79\% &50.83\%  & 91.88\%&\textbf{54.79\%}\\
    LLM-attacker & \textbf{63.54\%} & \textbf{45.00\%}&  \textbf{90.42\%}&\textbf{50.83\%} & \textbf{91.88\%}& \textbf{54.79\%}& \textbf{92.92\%} &\textbf{54.79\%}  & \textbf{93.54\%}&\textbf{54.79\%}\\ 
    \hline
    Computational Cost (s) & \multicolumn{2}{c|}{0.197} & \multicolumn{2}{c|}{0.263} & \multicolumn{2}{c|}{0.377} & \multicolumn{2}{c|}{0.450}  & \multicolumn{2}{c}{0.535}\\
    \hline
    \end{tabular}
    \label{tab:collision rate}
\end{table*}

\begin{table}[!htbp]
    \fontsize{9pt}{13pt}\selectfont
    \centering
    \caption{Realism of Driving Behavior}
    \begin{tabular}{c|cc}
    \hline
    Identification Method & \makecell{Acceleration \\ KL Divergence $\downarrow$} & \makecell{Abnormal Jerk \\ Rate $\downarrow$} \\
    \hline
        Random	 &0.0523 &	3.155\% \\ 
    Min TTC &	0.0134 &	1.961\% \\
    Kinetic Field	 &\textbf{0.0075}	 &1.864\% \\
    LLM-attacker	 &0.0112 &	\textbf{1.836\%} \\
    \hline
    \end{tabular}
    \label{tab:realism}
\end{table}

\begin{table*}[!htbp]
    \fontsize{8.5pt}{13pt}\selectfont
    \vspace{-10pt}
    \centering
    \caption{Adversarial Training Results}
    \begin{tabular}{c|cc|cc|cc}
    \hline
    \multirow{2}{*}{Training dataset} &  \multicolumn{2}{c}{Normal}  &  \multicolumn{2}{c}{One attacker} & \multicolumn{2}{c}{Two attackers} \\
    & Crash Rate $\downarrow$ & Route Completion $\uparrow$ & Crash Rate $\downarrow$ & Route Completion $\uparrow$ & Crash Rate $\downarrow$ & Route Completion $\uparrow$ \\
    \hline
    LLM-II &  \textbf{\textcolor{white}{0}8.81 $\pm$ 1.64\%}&	75.37 $\pm$ 3.41\%&	\textbf{18.00 $\pm$ 1.72\%}&	72.60 $\pm$\textcolor{white}{0}3.09\%&	\textbf{23.79 $\pm$ 1.32\%}&	70.92 $\pm$ 2.79\% \\ 
    LLM-I & 13.28 $\pm$ 1.89\%&	\textbf{78.61 $\pm$ 0.37\%}&	22.78 $\pm$ 1.63\%&	\textbf{75.16 $\pm$ 0.55\%}&	29.87 $\pm$ 1.15\%&	\textbf{72.82 $\pm$ 1.26\%}\\
    Replay & 22.61 $\pm$ 5.12\%&	74.48 $\pm$ 7.33\%&	40.25 $\pm$ 6.65\%&	68.52 $\pm$ 8.28\%&	45.07 $\pm$ 7.01\%&	65.80 $\pm$ 9.05\%\\
    Kinetic-II &19.50 $\pm$ 3.44\%&	61.93 $\pm$10.95\%&	31.10 $\pm$ 1.96\%&	56.91 $\pm$11.69\%&	37.36 $\pm$ 4.50\%&	53.82 $\pm$12.85\% \\
    \hline
    \end{tabular}
    \label{tab:rl training}
\end{table*}

Fig. \ref{fig:identification_example} demonstrates a specific example of identification results of different methods. 
This is an unprotected left-turn intersection scenario where the goal of ego vehicle is to turn left. The Min TTC identifies the vehicle from the right-side entrance lane as the attacker, and the Kinetic Field identifies the vehicle on the adjacent right lane of the ego vehicle. However, the two methods fail to attack the ego vehicle. In the LLM identification results, the oncoming vehicle from the opposite entrance lane is identified as the attacker, leading to a realistic and critical left-turn conflict. While the ego vehicle does not react in time, the collision between the ego vehicle and the attacker composes the safety-critical events.

\begin{figure}[htbp!]
    \centering
    \includegraphics[width=0.95\linewidth]{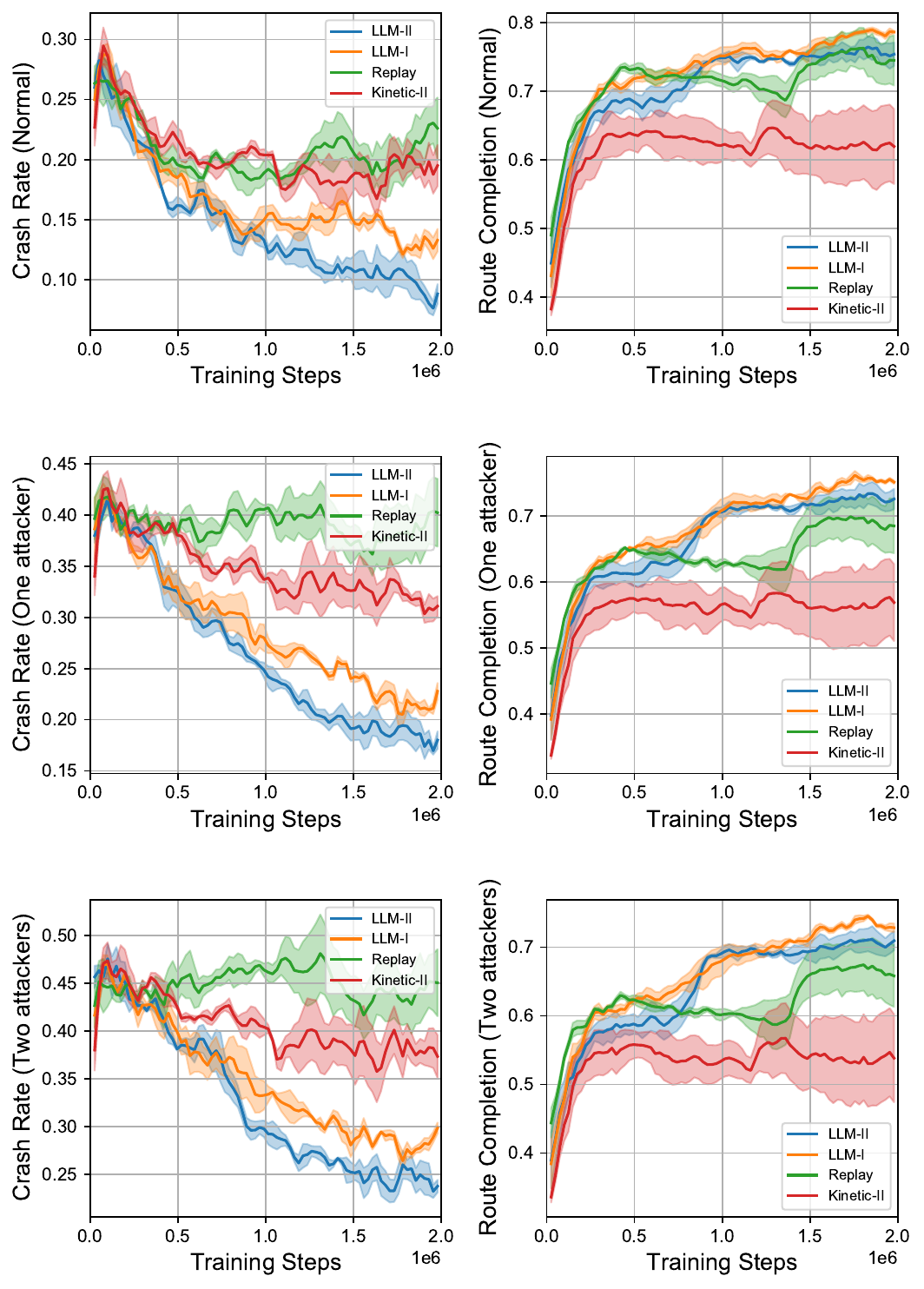}
    \caption{Adversarial training curves of the ADS policy. The X-axis represents the training steps and the Y-axis represents the testing metrics. Four types of training dataset (LLM-II, LLM-I, Replay and Kinetic-II) are compared, which are displayed in color blue, orange, green and red, respectively. The crash rate and route completion of the ADS policy are tested in real-world normal scenario, safety-critical scenario with one attacker, and safety-critical scenario with two attackers. The same set of scenarios was used across all tests, and the attackers, identified by the LLM, remained consistent throughout.}
    \label{fig:RL_training}
\end{figure}

\subsection{Adversarial training results}
In adversarial training, the generated safety-critical scenarios act as data augmentation during the training of ADS policy. The training dataset consists of two parts: 385 real-world WOMD replay scenarios and the safety-critical scenarios generated from these 385 scenarios. Three kinds of generation approaches are applied: (1) \textbf{LLM-II}: there are two attackers identified by LLM-attacker, (2) \textbf{LLM-I}: only one attacker identified by LLM-attacker, (3) \textbf{Kinetic-II}: two attackers identified by Kinetic Field. Meanwhile, training based on the \textbf{Replay} dataset, which consists of only real-world replay scenarios, is also conducted for comparison. The dataset includes the other 100 real-world normal replay scenarios, and their generated safety-critical scenarios (with one attacker and two attackers) are utilized for testing. In the testing, two metrics are evaluated: (1) the crash rate, which means the proportion of collisions between the ADS and other vehicles in the test scenario, and the lower the better. (2) the route completion, which represents the driving task completion rate of ADS and the higher the better for ADS. To ensure consistency, the four types of training dataset are repeated three times using the same random seed. Then, mean of the test metrics and half of variance of the test metrics are calculated.

The adversarial training results are shown in Fig. \ref{fig:RL_training} and TABLE \ref{tab:rl training}.  For the crash rate, the policy trained with the LLM-II achieves the lowest collision rate, which reduces collisions by approximately half compared to Replay training in both normal and safety-critical scenarios. For the route completion, the policy trained with LLM-I performs the best while the policy trained with LLM-II ranks second. This may be attributed to the stronger adversarial pressure introduced by LLM-II with two attackers, which encourages the policy to adopt more conservative behaviors. In general, it can be demonstrated that the safety-critical scenario with one or two attackers identified by LLM outperforms the Replay scenarios or the Kinetic identification. The results prove that enhancing the training dataset by effective safety-critical scenarios can greatly improve the safety and robustness of the ADS policy.

\section{Conclusion}\label{sec:conclusion}
Safety-critical scenarios are rare in real world and adversarial scenario generation methods, which modify the trajectories in real-world traffic scenario, can turn normal scenarios into safety-critical ones. This paper presents LLM-attacker: a closed-loop adversarial scenario generation framework leveraging LLMs to address the adversarial vehicle identification and closed-loop training problem. Multiple collaborative LLM agents are manipulated to realize automatic adversarial vehicle identification. The generated scenarios are applied to the testing and training of ADS. It turns out that the attack success rate of the LLM-attacker outperforms other methods, and two attackers achieve the balance between attack effectiveness and computational efficiency in the testing process. Furthermore, the safety performance of the ADS trained by LLM-attacker is significantly better than that trained by normal scenarios.

There are several potential directions to be explored in the future. First, newly released LLMs can be compared to assess their ability in adversarial scenario generation, including reasoning efficiency and adversarial effectiveness. Second, incorporating traffic rule constraints into attacker identification and trajectory generation may generate more diverse and edge-case scenarios.

\bibliographystyle{IEEEtran}
\bibliography{ref-extracts}
\vspace{-10ex} 

\begin{IEEEbiography}[{\includegraphics[width=1in,height=1.1in,keepaspectratio]{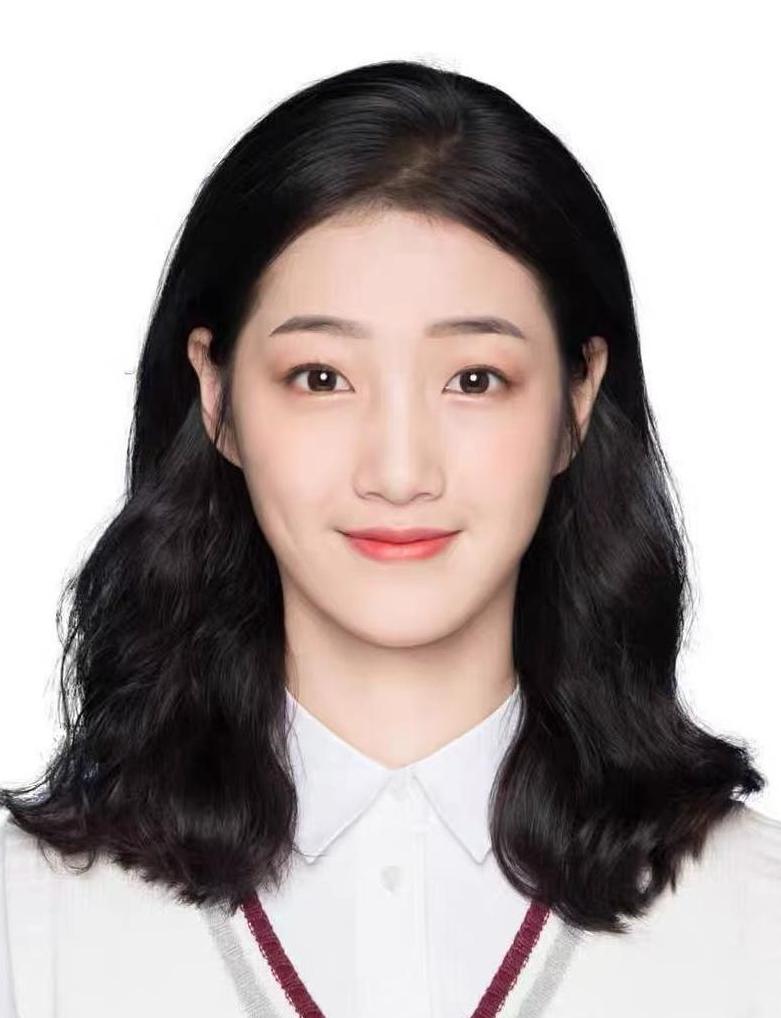}}]{Yuewen Mei} received the B.S. degree in transportation engineering from Central South University, Changsha, Hunan, China. Now she is currently pursuing the Ph.D. degree with the Department of Traffic Engineering in Tongji University, Shanghai, China. Her research interests include fault injection safety testing and safety-critical scenario generation for highly automated vehicles.
\end{IEEEbiography}

\vspace{-10ex} 

\begin{IEEEbiography}[{\includegraphics[width=1in,height=1.1in,keepaspectratio]{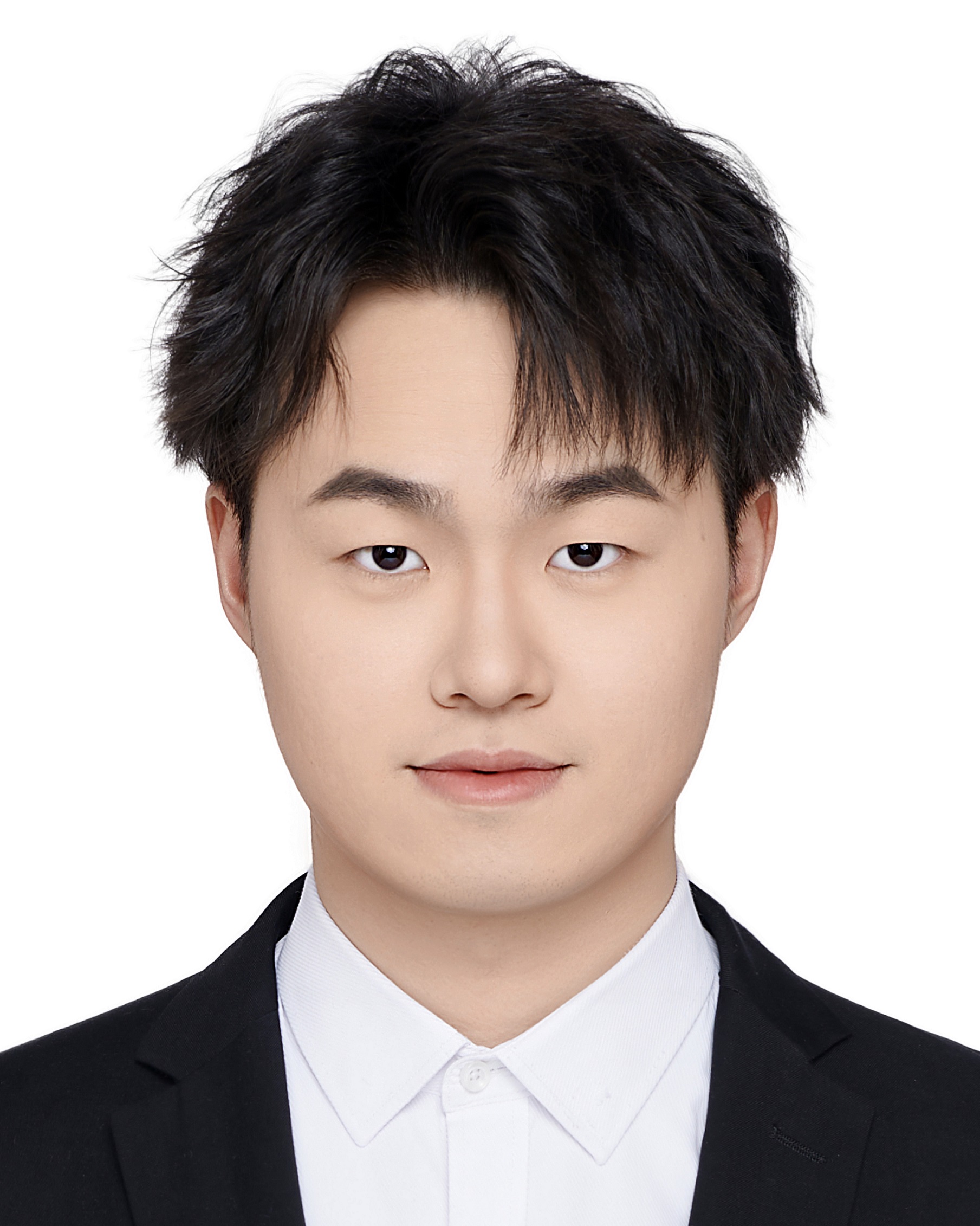}}]{Tong Nie} received the B.S. degree from Tongji University, Shanghai, China. He is currently pursuing dual Ph.D. degrees with Tongji University and The Hong Kong Polytechnic University. He has published several papers in top-tier journals and conferences in AI and transportation, such as KDD, AAAI, IEEE TITS, and TR-Part C/E.
His research interests include large language models and autonomous driving. His research was supported by a national grant.
\end{IEEEbiography}
\vspace{-10ex} 

\begin{IEEEbiography}[{\includegraphics[width=1in,height=1.1in,keepaspectratio]{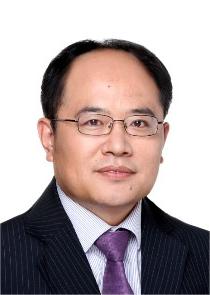}}]{Jian Sun}, received the Ph.D. degree in transportation engineering from Tongji University, Shanghai, China. He is currently a Professor of transportation engineering with Tongji University. His research interests include intelligent transportation systems, traffic flow theory, AI in transportation, and traffic simulation. 
\end{IEEEbiography}

\vspace{-10ex} 

\begin{IEEEbiography}[{\includegraphics[width=1in,height=1.1in,keepaspectratio]{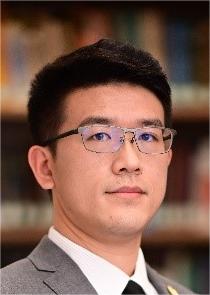}}]{Ye Tian (Senior Member, IEEE)} he received the Ph.D. degree in transportation engineering from The University of Arizona, Tucson, AZ, USA, in 2015. He is currently a Professor of transportation engineering with Tongji University, Shanghai, China. He serves as an Associate Editor of IEEE Transactions of Intelligent Transportation Systems. His research interests include active demand management, dynamic traffic assignment, mesoscopic traffic simulation, and safety assurance of automated vehicles.
\end{IEEEbiography}

\end{document}